\documentclass[letterpaper, 10 pt, conference]{ieeeconf}  % Comment this line out if you need a4paper

\IEEEoverridecommandlockouts
\makeatletter
\let\NAT@parse\undefined
\makeatother
\usepackage{svg}
\usepackage{hyperref}
\hypersetup{
    colorlinks,
    filecolor=black,
    linkcolor=black,
    urlcolor=black
}
\usepackage{url}
\usepackage{multirow}
\usepackage{amsmath}
\usepackage{tikz}
\usetikzlibrary{arrows.meta}
\usepackage[labelsep=period]{caption}
\usepackage{nicefrac}
\usepackage{stmaryrd}

\usepackage{todonotes}
\begin{document}

\title{\LARGE \bf
Multi range Real-time depth inference from a monocular stabilized footage using a Fully Convolutional Neural Network
}

\author{Cl{\'e}ment Pinard$^{a,b}$ , Laure Chevalley$^{a}$, Antoine Manzanera$^{b}$, David Filliat$^{b}$% <-this % stops a space
\thanks{$^{a}$Parrot, Paris, France }
\thanks{\texttt{(clement.pinard, laure.chevalley)@parrot.com}}
\thanks{$^{b}$U2IS, ENSTA ParisTech,
Universit\'e Paris-Saclay, Palaiseau, France}
\thanks{\texttt{(clement.pinard, antoine.manzanera, david.filliat)@ensta-paristech.fr}}%
}

\maketitle
\begin{abstract}
Using a neural network architecture for depth map inference from monocular stabilized videos with application to UAV videos in rigid scenes, we propose a multi-range architecture for unconstrained UAV flight, leveraging flight data from sensors to make accurate depth maps for uncluttered outdoor environment.

We try our algorithm on both synthetic scenes and real UAV flight data. Quantitative results are given for synthetic scenes with a slightly noisy orientation, and show that our multi-range architecture improves depth inference.

Along with this article is a video that present our results more thoroughly.
\end{abstract}

\IEEEpeerreviewmaketitle

\section{Introduction}
Scene understanding from vision is a core problem for autonomous vehicles and for UAVs in particular.
In this paper we are specifically interested in computing the depth of each pixel from image sequences captured by a camera. We assume our camera's velocity (and thus displacement between two frames) is known, as most UAV flight systems include a speed estimator, allowing to settle the scale invariance ambiguity of the depth map.

Solving this problem could be beneficial for several problems such as environment scanning or applying depth-based sense and avoid algorithms for lightweight embedded systems that only have a monocular camera. 
Not relying on depth Sensors such as stereo vision, ToF camera, LiDar or Infra Red emitter/receiver allows to free the UAV from their weight, cost and limitations. Specifically, along with some RGB-D sensors being unable to operate under sunlight (e.g. IR and ToF), most of them suffer from range limitations and can be inefficient in case we need long-range information such as trajectory planning \cite{hadsell2009learning}.
Unlike RGB-D sensors, depth from motion is flexible w.r.t. displacement and thus robust to high speeds or high distances as choosing among previous frames gives us a wide range of different displacements.
For estimating such depth maps, we designed an end-to-end learning architecture, based on a synthetic dataset and a fully convolutional neural network that takes as input an image pair taken at different times. No preprocessing such as optical flow computation, nor visual odometry is applied to the input, while the depth is directly provided as an output. \cite{Pinard_uavg}

%The dataset we created is not providing totally unconstrained movement between image pairs. 
We created a dataset of image pairs with random translation movements, with no rotation, and a constant displacement magnitude applied during the whole training.
\begin{figure}
\centering
\begin{tabular}{rll}
a)\includegraphics[scale=0.06]{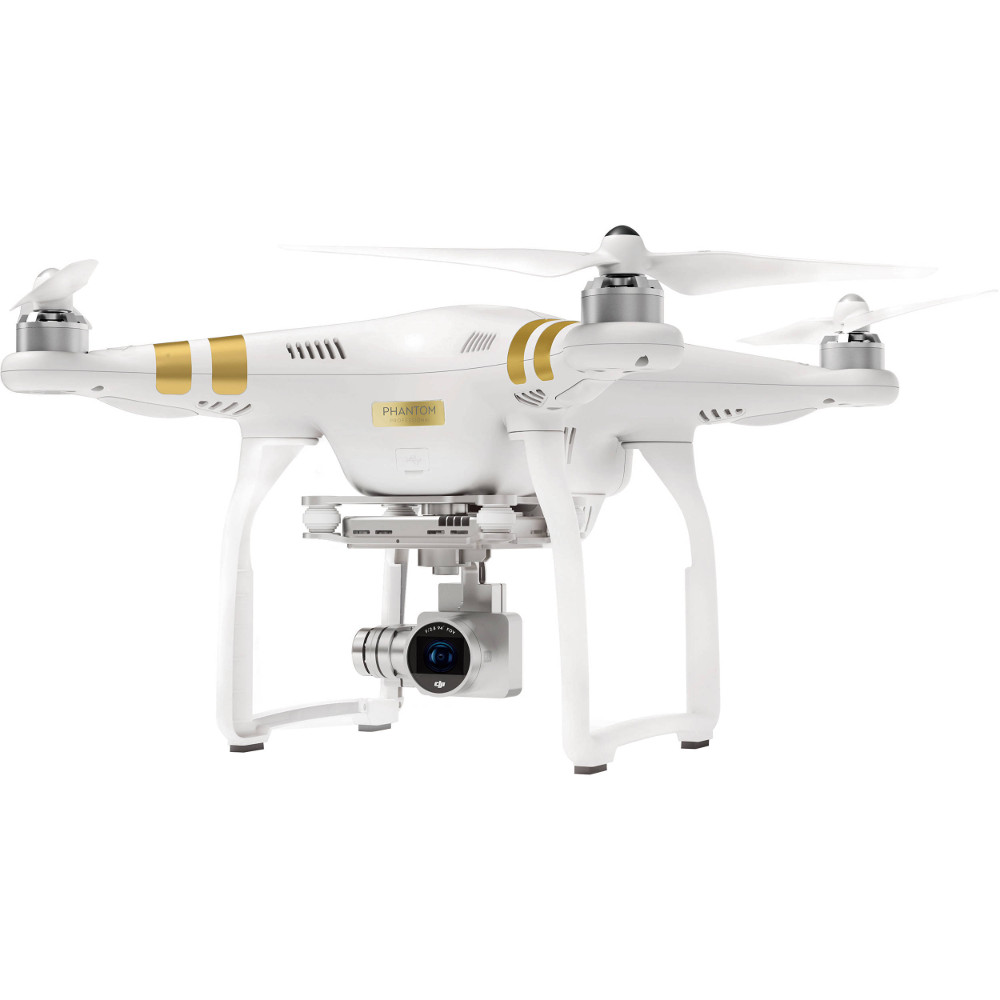} &
b)\includegraphics[scale=0.06]{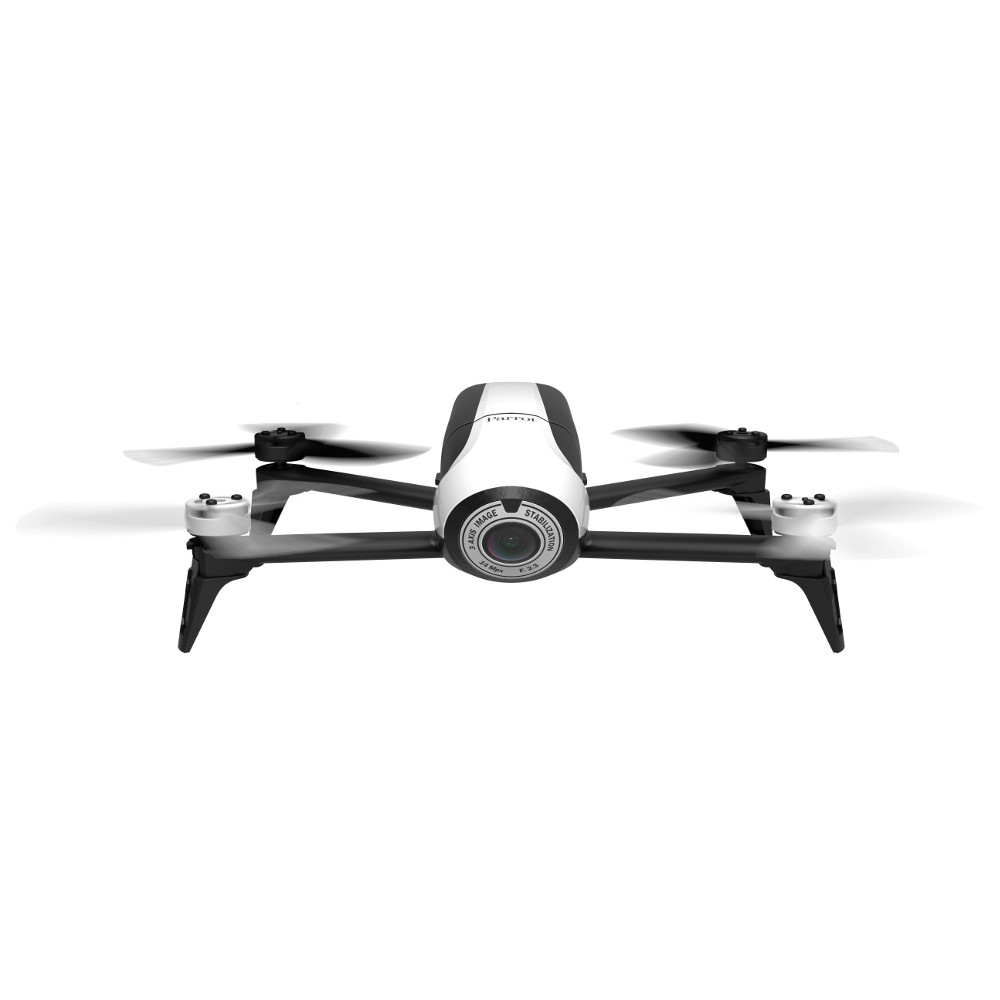} &
c)\includegraphics[scale=0.1]{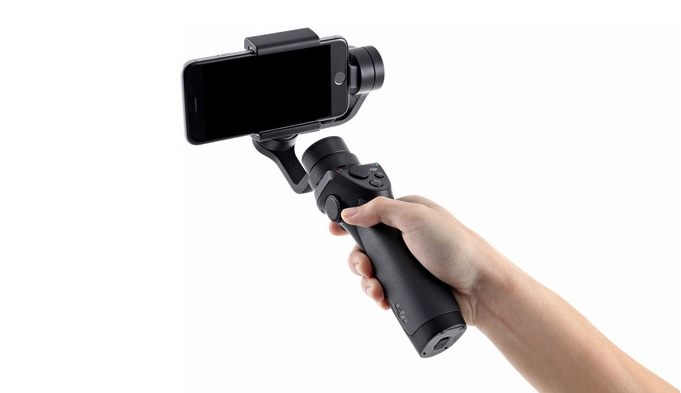}
\end{tabular}
\caption{Camera stabilization can be done via a) mechanic gimbal or b) dynamic cropping from fish-eye camera, for drones or c) hand-held cameras}
\label{drones}
\end{figure}

The assumption about videos without rotation appears realistic for two reasons:
\begin{itemize}
\setlength\itemsep{0em}\setlength\parskip{0em}\setlength\topsep{0em}\setlength\partopsep{0em}\setlength\parsep{0em}
\item Hardware rotation compensation is 
mainly a solved problem, even for consumer products, with IMU-stabilized cameras on consumer drones or hand-held steady-cam (Fig~\ref{drones}).
\item this movement is somewhat related to human vision and vestibulo-ocular reflex (VOR) \cite{VOR}. Our eyes orientation is not induced by head rotation, our inner ear among other biological sensors allows us to compensate parasite rotation when looking at a particular direction.
\end{itemize}

Using the trained network, we propose an algorithm for real condition depth inference from a stabilized UAV. Displacement from sensors is used to compute real depth map, as it only differs from the synthetic constant displacement images by a scale factor. Our network output also allows us to {\it a posteriori} optimize the depth inference. By adjusting frame shift to get a displacement that would make the network get the same disparity distribution as during its training, we lower the depth error for next inference. For example, with large distances, ideal displacement between two frames is higher, and thus the shift is also higher for a given speed.
 Moreover, we use multiple batch inference to compute multiple depth maps centered around a particular range, and fuse them to get a high precision for both close and far objects, no matter the distance, given a sufficient displacement from the UAV.

\section{Related Work}

Deep Learning and Convolutional Neural Networks have recently been widely used for numerous kinds of vision problem such as classification \cite{krizhevsky2012imagenet} and hand-written digits recognition \cite{lecun1998gradient}.

Depth from vision is one of the problems studied with neural network, and has been addressed with a wide range of training solution. Some datasets \cite{geiger2012we,Silberman:ECCV12} allow a neural network to learn end-to-end depth or disparity \cite{luo2016efficient,zbontar2015computing,eigen2014depth}.
Reprojection error has also been used for unsupervised training for depth from a single image \cite{2017arXiv170407804V,zhou2017unsupervised} or for disparity between two frames of a stereo rig \cite{DBLP:journals/corr/KondaM13,DBLP:journals/corr/GargBR16}.

Depth from a single image, although interesting, suffers from a major drawback which is overfitting. No motion is given to the network during inference, and the resulting depth is inferred from context, whereas they can be decorrelated. This technique can be sufficient for road driving context with an obvious road in front of the camera, but for a UAV flight usage, we may have to deal with very heterogeneous scenes. On the other hand, depth from a stereo pair is only implying a single lateral movement, and lacks a forward component to appear realistic for any aerial stabilized footage.

For depth from more complex movement from a monocular camera, current state of the art methods tend to use motion, and especially structure from motion, and most algorithm do not rely on deep learning \cite{cadena2016past,mur2016orb,klein2007parallel}. Prior knowledge w.r.t. scene is used to infer a sparse depth map with its density usually growing over time.
These techniques also called SLAM are typically used with unstructured movement (translation and rotation with varying magnitudes), produce very sparse point-cloud based 3D maps and require heavy calculation to keep track of the scene structure and align newly detected 3D points to the existing ones.

Our goal is to compute a dense depth map (where every point has a valid depth) using only two frames from the same camera, at different times, and without prior knowledge on the scene and direction of movement, apart from the lack of rotation and the scale factor.

\section{End-to-end learning of Depth Inference}
\label{depth_part}
Inspired by flow estimation and disparity (which is essentially magnitude of optical flow vectors), a problem to which exist a lot of very convincing methods \cite{ilg2016flownet,2017arXiv170304309K}, we set up an end-to-end learning workflow, by training a neural network to explicitly predict the depth of every pixel in a scene, from an image pair with constant displacement value.

\subsection{Still Box Dataset}

\begin{figure}
\centering
\begin{tabular}{ll}
\includegraphics[scale=0.2]{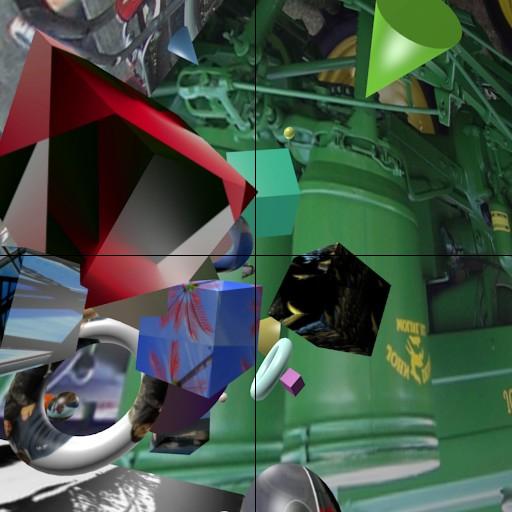} & \includegraphics[scale=0.2]{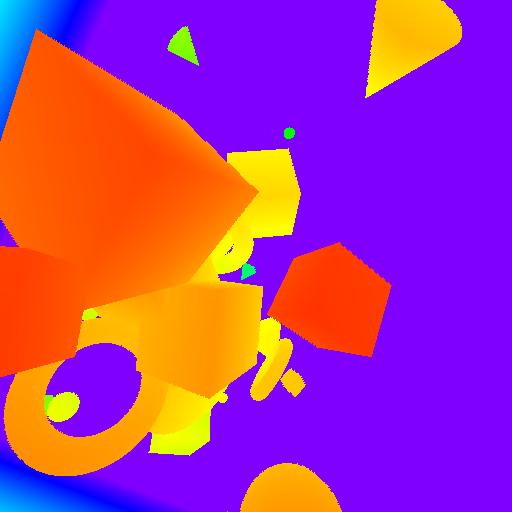} \\
\includegraphics[scale=0.2]{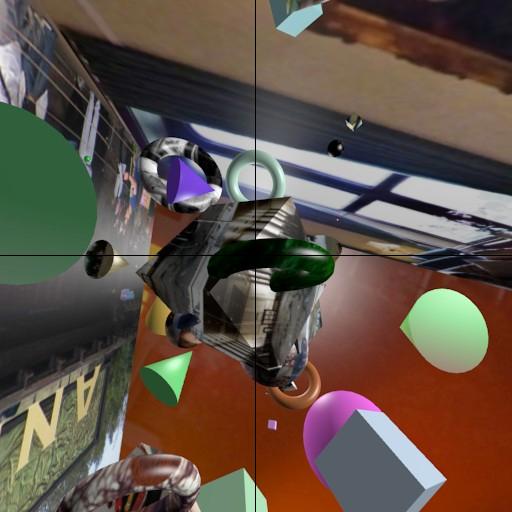} & \includegraphics[scale=0.2]{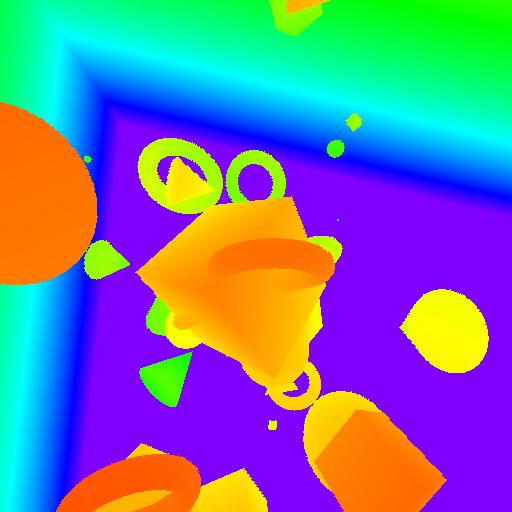}
  
\end{tabular}
\caption{Some examples of our renderings with associated depth maps (red is close, purple is far)}
\label{pictures}
\end{figure}

We design our own synthetic dataset, using the rendering software {\it Blender}, to generate an arbitrary number of random rigid scenes, composed of basic 3d primitives (cubes, spheres, cones and tores) randomly textured from an image set scrapped from {\it Flickr} (see Fig~\ref{pictures}).

These objects are randomly placed and sized in the scene, and walls are added at large distances as if the camera was inside a box (hence the name). The camera is moving at a fixed speed value, but to an uniformly distributed random direction, which is constant for each scene. It can be anything from forward/backward movement to lateral movement (which is then equivalent to stereo vision).

\subsection{Dataset augmentation}

In our dataset, we store data in 10 images long videos, with each frame paired with its ground truth depth. This allows us to set {\it a posteriori} distances distribution with a variable temporal shift between two frames. 
If we use a baseline shift of 3 frames, we can e.g. assume a depth three times as great as for two consecutive frames (shift of 1). 
In addition, we can also consider negative shift, which will only change displacement direction without changing speed value. This allows us, given a fixed dataset size, to get more evenly distributed depth values to learn, and also to de-correlate images from depth, preventing over-fitting during training, that would result in a scene recognition algorithm and would poorly perform on a validation set.

\subsection{Depth Inference training}

\begin{figure}
\begin{tabular}{|l|}
  \hline
  Typical Conv Module \\ \hline \hline
  SpatialConv, 3x3\\ \hline
  SpatialBatchNorm\\ \hline
  ReLU\\ \hline
  
\end{tabular}
\begin{tabular}{|l|}
  \hline
  Typical Deconv Module \\ \hline \hline
  SpatialConvTranspose, 4x4\\ \hline
  SpatialConv, 3x3\\ \hline
  SpatialBatchNorm\\ \hline
  ReLU\\ \hline
  
\end{tabular}

\tikzset{%
  >={Latex[width=2mm,length=2mm]},
  % Specifications for style of nodes:
            base/.style = {rectangle, rounded corners,draw=black,
                           minimum width=2.2cm},
            Conv/.style = {base, fill=blue!30},
            Deconv/.style = {base, fill=red!30},
            Concat/.style = {base, fill=green!30},
            Depth/.style = {base, fill=orange!15},
            Loss/.style = {base, fill=purple!30},
}

\begin{tikzpicture}[scale = 0.4,font=\sffamily\small]
  % Specification of nodes (position, etc.)
  \node (start)             [base]              {Input image pair};
  \node (Conv1)      [Conv, below of=start]          {Conv1, stride 2};
  \node (Conv2)      [Conv, below of=Conv1]   {Conv2, stride 2};
  \node (Conv3)      [Conv, below of=Conv2]   {Conv3, stride 2};
  \node (Conv3_1)    [Conv, below of=Conv3]   {Conv3.1};
  \node (Conv4)      [Conv, below of=Conv3_1]   {Conv4, stride 2};
  \node (Conv4_1)    [Conv, below of=Conv4]   {Conv4.1};
  \node (Conv5)      [Conv, below of=Conv4_1]   {Conv5, stride 2};
  \node (Conv5_1)    [Conv, below of=Conv5]   {Conv5.1};
  \node (Conv6)      [Conv, below of=Conv5_1]   {Conv6, stride 2};
  \node (Conv6_1)    [Conv, below of=Conv6]   {Conv6.1};
  \node (Deconv5)    [Deconv, below of=Conv6_1] {Deconv5};
  \node (Concat5)    [Concat, below of=Deconv5] {Concat5};
  \node (Deconv4)    [Deconv, below of=Concat5] {Deconv4};
  \node (Concat4)    [Concat, below of=Deconv4] {Concat4};
  \node (Deconv3)    [Deconv, below of=Concat4] {Deconv3};
  \node (Concat3)    [Concat, below of=Deconv3] {Concat3};
  \node (Deconv2)    [Deconv, below of=Concat3] {Deconv2};
  \node (Concat2)    [Concat, below of=Deconv2] {Concat2};
  \node (Depth2)     [Depth, below of=Concat2] {Depth2};
  \node (finish)      [base, below of=Depth2] {Final depth output};
  \node (L1Loss)     [Loss, right of=Depth2, xshift=4cm, yshift=-1cm,text width=2.2cm, align=center] {MultiScale L1 Loss};
  
  \node (Depth6)     [Depth, right of=Conv6_1, xshift=2.5cm, yshift= -0.5cm] {Depth6};
  \node (UpDepth6)    [Deconv, below of=Depth6] {Up Depth6};
  
  \node (Depth5)     [Depth, right of=Concat5, xshift=2.5cm, yshift= -0.5cm] {Depth5};
  \node (UpDepth5)    [Deconv, below of=Depth5] {Up Depth5};
  
  \node (Depth4)     [Depth, right of=Concat4, xshift=2.5cm, yshift= -0.5cm] {Depth4};
  \node (UpDepth4)    [Deconv, below of=Depth4] {Up Depth4};
  
  \node (Depth3)     [Depth, right of=Concat3, xshift=2.5cm, yshift= -0.5cm] {Depth3};
  \node (UpDepth3)    [Deconv, below of=Depth3] {Up Depth3};
     
  % Specification of lines between nodes specified above
  % with aditional nodes for description 
  \draw[->]             (start) -- (Conv1) node [midway,right] {6xHxW};
  \draw[->]     (Conv1) -- (Conv2) node [midway,right] {32x\nicefrac{1}{2}Hx\nicefrac{1}{2}W};
  \draw[->]     (Conv2) -- (Conv3)node [midway,right] {64x\nicefrac{1}{4}Hx\nicefrac{1}{4}W};
  \draw[->]     (Conv3) -- (Conv3_1)node [midway,right] {128x\nicefrac{1}{8}Hx\nicefrac{1}{8}W};
  \draw[->]     (Conv3_1) -- (Conv4);
  \draw[->]     (Conv4) -- (Conv4_1)node [midway,right] {256x\nicefrac{1}{16}Hx\nicefrac{1}{16}W};
  \draw[->]     (Conv4_1) -- (Conv5);
  \draw[->]     (Conv5) -- (Conv5_1)node [midway,right] {256x\nicefrac{1}{32}Hx\nicefrac{1}{32}W};
  \draw[->]     (Conv5_1) -- (Conv6);
  \draw[->]     (Conv6) -- (Conv6_1)node [midway,right] {512x\nicefrac{1}{64}Hx\nicefrac{1}{64}W};
  \draw[->]     (Conv6_1) -- (Deconv5);
  \draw[->]     (Conv5_1.west) -- +(-0.6,0) |- (Concat5);
  \draw[->]     (Deconv5) -- (Concat5)node [midway,right] {256x\nicefrac{1}{32}Hx\nicefrac{1}{32}W};
  \draw[->]     (UpDepth6) |- (Concat5);
  \draw[->]     (Concat5) -- (Deconv4);
  \draw[->]     (Deconv4) -- (Concat4)node [midway,right] {128x\nicefrac{1}{16}Hx\nicefrac{1}{16}W};
  \draw[->]     (Conv4_1.west) -- +(-0.8,0) |- (Concat4);
  \draw[->]     (UpDepth5) |- (Concat4);
  \draw[->]     (Concat4) -- (Deconv3);
  \draw[->]     (UpDepth4) |- (Concat3);
  \draw[->]     (Conv3_1.west) -- +(-1,0) |- (Concat3);
  \draw[->]     (Deconv3) -- (Concat3)node [midway,right] {64x\nicefrac{1}{8}Hx\nicefrac{1}{8}W};
  \draw[->]     (Concat3) -- (Deconv2);
  \draw[->]     (Conv2.west) -- +(-1.2,0) |- (Concat2);
  \draw[->]     (Deconv2) -- (Concat2)node [midway,right] {32x\nicefrac{1}{4}Hx\nicefrac{1}{4}W};
  \draw[->]     (UpDepth3) |- (Concat2);
  \draw[->]     (Concat2) -- (Depth2);
  \draw[->]     (Conv6_1) |- (Depth6);
  \draw[->]     (Concat5) |- (Depth5);
  \draw[->]     (Concat4) |- (Depth4);
  \draw[->]     (Concat3) |- (Depth3);
  \draw[->]     (Depth6) -- (UpDepth6)node [midway,right] {1x\nicefrac{1}{32}Hx\nicefrac{1}{32}W};
  \draw[->]     (Depth5) -- (UpDepth5)node [midway,right] {1x\nicefrac{1}{16}Hx\nicefrac{1}{16}W};
  \draw[->]     (Depth4) -- (UpDepth4)node [midway,right] {1x\nicefrac{1}{16}Hx\nicefrac{1}{16}W};
  \draw[->]     (Depth3) -- (UpDepth3)node [midway,right] {1x\nicefrac{1}{8}Hx\nicefrac{1}{8}W};
  \draw[->]     (Depth6) -| (L1Loss.30);
  \draw[->]     (Depth5) -| (L1Loss.30);
  \draw[->]     (Depth4) -| (L1Loss.30);
  \draw[->]     (Depth3) -| (L1Loss.30);
  \draw[->]     (Depth2) -| (L1Loss.30);
  \draw[->]     (Depth2) -- (finish) node [midway,right] {1x\nicefrac{1}{4}Hx\nicefrac{1}{4}W};
  \end{tikzpicture}
\linebreak
\caption{DepthNet structure parameters, Conv and Deconv modules detailed above}
\label{network}
\end{figure}
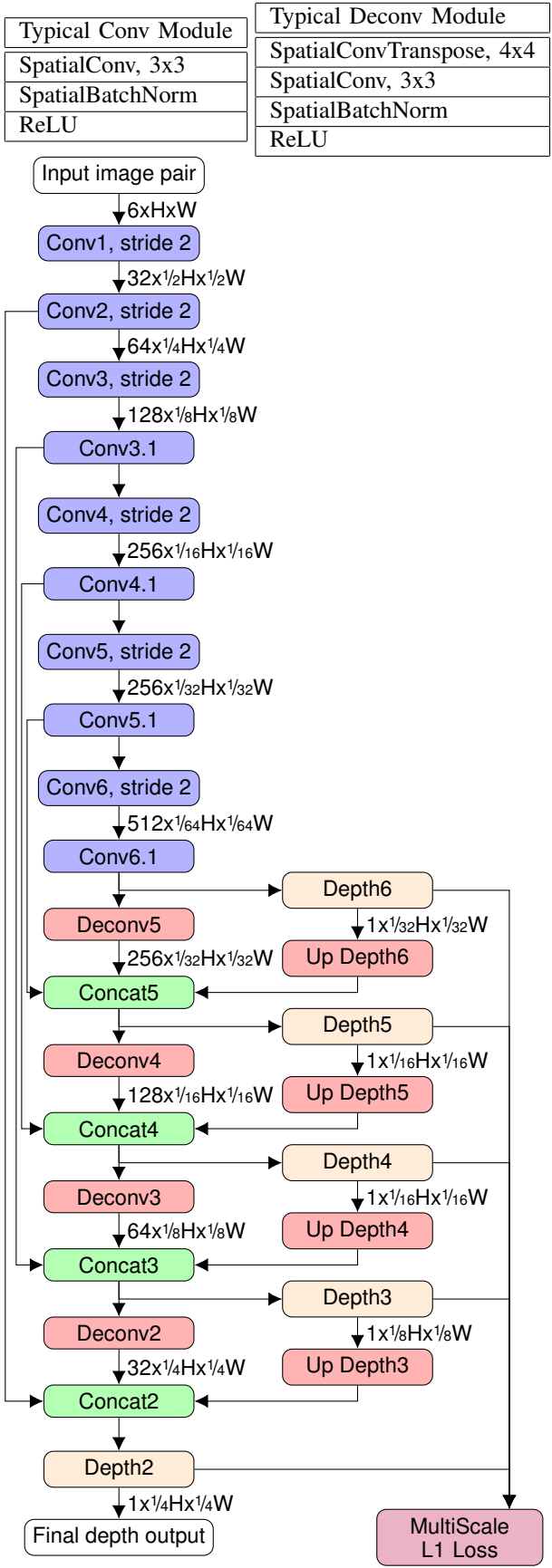

Our network is broadly inspired from FlowNetS \cite{DFIB15} (initially used for flow inference) and called DepthNet. It is described in details in \cite{Pinard_uavg}, we provide here a summary of its structure (Fig~\ref{network}) and performances. Each convolution (apart from depth modules) is followed by a Spatial Batch Normalization and ReLU activation layer. Batch normalization helps convergence and stability during training by normalizing a convolution's output (0 mean and standard deviation of 1) over a batch of multiple inputs \cite{ioffe2015batch}, and Rectified Linear Unit (ReLU) is the typical activation layer \cite{DBLP:journals/corr/XuWCL15}. Depth Module are convolution modules, reducing the input to $1$ feature map, which is expected to be the depth map, at a given scale. One should note that FlowNetS initially used LeakyReLU which has a non-null slope for negative values, but tests showed that ReLU performed better for our problem.

The main idea behind this network is that upsampled feature maps are concatenated with corresponding earlier convolution outputs (e.g. Conv2 output with Deconv5 output). Higher semantic information is then associated with information more closely linked to pixels (since it went through less downsampling convolutions) which is then used for reconstruction.

This multi-scale architecture has been proven very efficient for flow and disparity computing while keeping a very simple supervised learning process.
%Our network is admittedly very simple and one could leverage some more advanced work that were used for flow and disparity, such as FlowNetC or GC-Net \cite{2017arXiv170304309K} among many others.

The main point of this experimentation is to show that direct depth estimation can be efficient regarding unknown translation direction. Like FlowNetS, we use a multi-scale criterion, with a L1 reconstruction error for each scale:

\begin{equation}
Loss = \sum_{s\in scales} \gamma_s\frac{1}{H_sW_s} \sum_i \sum_j \left| \beta_s(i,j) - \zeta_s(i,j)\right|
\end{equation}
where
\begin{itemize}
\setlength\itemsep{0em}\setlength\parskip{0em}\setlength\topsep{0em}\setlength\partopsep{0em}\setlength\parsep{0em}
\item $\gamma_s$ is the weight of the scale, arbitrarily chosen.
%\item $(H_s,W_s) = (\nicefrac{1}{2^n}H,\nicefrac{1}{2^n}W)$
\item $(H_s,W_s) = (\nicefrac{1}{2^s}H,\nicefrac{1}{2^s}W)$
are the height and width of the output.
\item $\zeta_s$ is the scaled depth groundtruth, using average pooling.
\item $\beta_s$ is the ouput of the network at scale $s$.
\end{itemize}

As said earlier, we apply data augmentation to the dataset using different shifts, along with classic methods such a flips and rotations. We also clip depth to a maximum of 100m, and provide sample pairs without shift, assuming its depth is 100m everywhere. As a consequence, the trained network will only be able to infer depth lower than 100m.

\begin{figure}\centering
\includegraphics[scale=0.2]{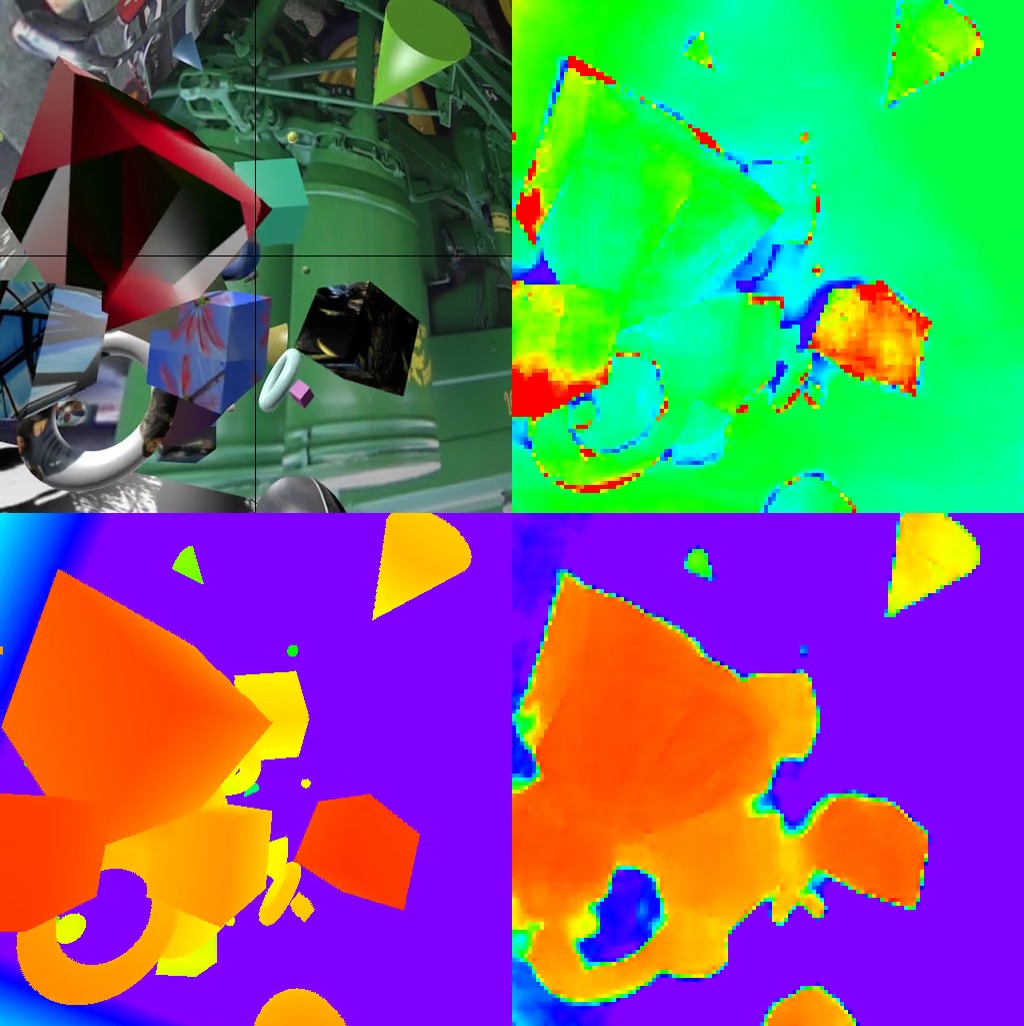}
\caption{Result on 512x512 images from DepthNet$_{64\rightarrow128\rightarrow256\rightarrow512}$. Upper-left: input, lower-left: Ground Truth depth, lower-right: our network output (128x128), upper-right: error, green is no error, red is overestimated depth, blue is underestimated}
\label{result512}
\end{figure}

\begin{table}
\centering
\begin{tabular}{|l|c|c|c|c|c}
  \hline
  \multirow{2}{*}{Network} & \multicolumn{2}{|c|}{L1Error} & \multicolumn{2}{|c|}{RMSE}\\ \cline{2-5}
  &train&test&train&test\\ \hline
  FlowNetS$_{64}$&$1.69$&$4.16$&$4.25$&$7.97$\\ \hline
  DepthNet$_{64}$&$2.26$&$4.49$&$5.55$&$8.44$		\\ \hline
  FlowNetS$_{64\rightarrow128\rightarrow256\rightarrow512}$&$0.658$&$\textbf{2.44}$&$1.99$&$\textbf{4.77}$\\ \hline
  DepthNet$_{64\rightarrow128}$&$1.20$&$3.07$&$3.43$&$6.30$\\ \hline
  DepthNet$_{64\rightarrow128\rightarrow256}$&$0.876$&$\textbf{2.44}$&$2.69$&$4.99$\\ \hline
  DepthNet$_{64\rightarrow128\rightarrow256\rightarrow512}$&$1.09$&$2.48$&$2.86$&$\textbf{4.90}$\\ \hline
  DepthNet$_{64\rightarrow512}$&$1.02$&$2.57$&$2.81$&$5.13$\\ \hline
  DepthNet$_{512}$&$1.74$&$4.59$&$4.91$&$8.62$\\ \hline
  
\end{tabular}
\caption{Quantitative results for depth inference networks. FlowNetS is modified with 1 channel outputs (instead of 2 for flow), trained from scratch for depth with Still Box, subscript indicates fine tuning process.}
\label{quantitative}
\end{table}

We applied training on several input size images, from 64x64 to 512x512. Fig~\ref{result512} shows training results for mean L1 reconstruction error. Like FlowNetS, network output are downsampled by a factor of 4 with reference to the input size.  As Table~\ref{quantitative} shows, best results are obtained with multiple fine-tuning, with intermediate scales $64$, $128$, $256$,  and finally $512$ pixels. Subscript values indicate finetuning processes. FlowNetS is performing better than DepthNet but by a fairly light margin while being 5 times heavier and most of the time much slower.

\section{UAV navigation use-case}
\label{uav}

\begin{figure} \centering
\includegraphics[scale=0.2]{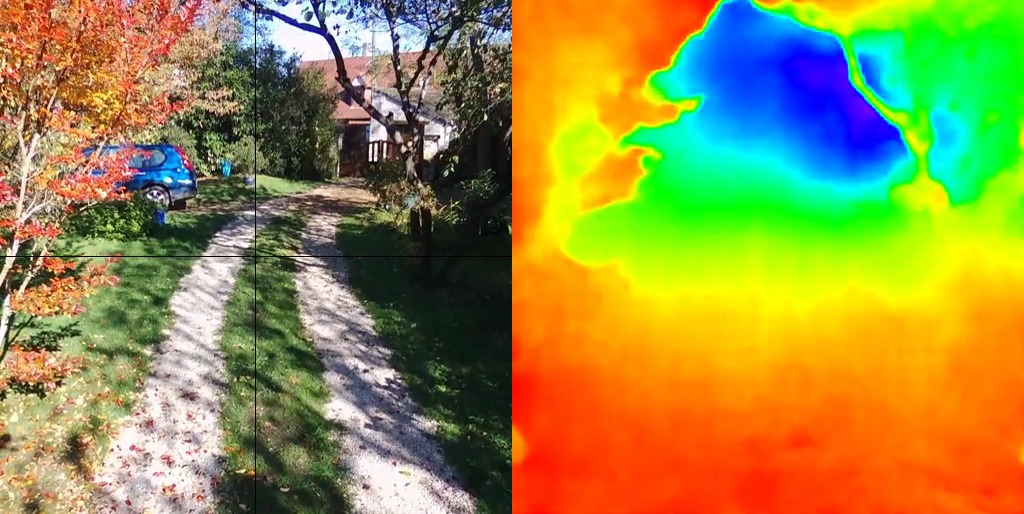}
\caption{Result on $512$x$512$ real images input from a Bebop drone footage}
\label{resultirl}
\end{figure}

\subsection{Optimal frame shift determination}
We learned depth inference from a moving camera, assuming its velocity is always the same. Results from real condition drone footage, on which we were careful to avoid camera rotation can be seen Fig~\ref{resultirl}. These results did not benefit from any fine-tuning from real footage, indicating that our Still Box Dataset, although not realistic in its scenes structures and rendering, appears to be sufficiently heterogeneous for learning to produce decent depth maps in real conditions. When running during flight, such a system can deduce the real depth map $\zeta$ from the network output and the drone displacement, knowing that the training displacement was $D_0$ (here $0.3m$)

\begin{equation}
\begin{array}{l}
\zeta(t) = DepthNet(I_t,I_{t-\Delta t})\frac{D(t,\Delta t)}{D_0}\\
\\
D(t,\Delta t) = \left\Vert\int_{t-\Delta t}^{t}V(\tau)d\tau\right\Vert
\end{array}
\end{equation}

The actual correct interpretation of the output of DepthNet is rather a percentage than a distance. $100\%$ meaning max distance for a given displacement $D$. We can introduce a function $\beta = \frac{DepthNet(I_t,I_{t-\Delta t})}{maxDistance}$ and a dimension-less parameter $\alpha = \frac{maxDistance}{D_0}$ for computing actual depth using the displacement $D$ as the only distance related factor.

\begin{equation}
\zeta(t) = \alpha \beta(I_t,I_{t-\Delta t})D(t,\Delta t)
\end{equation}

Depending of the depth distribution of the ground-truth depth map, it may be useful to adjust frame shift $\Delta t$. For example, when flying high above the ground with low speed, big structure detection and avoidance requires knowing precise distance values that are outside the typical range of any RGB-D sensor. The logical strategy would then be to increase the temporal shift between the frame pairs provided to DepthNet as inputs. More generally, one must provide inputs to DepthNet in order to ensure a well distributed depth output within its typical range. Depth-wise normalized error which is the essential quality measurement for values that we want to rescale, will diverge when ground truth depth approaches $0$. Indeed, in addition to being equivalent to an infinite optical flow, the depth-wise error cannot tend to $0$, which will make the expression $\nicefrac{error}{depth}$ tend to $+\infty$ at $0$
We thus need to choose the optimal spatial displacement and corresponding temporal shift to minimize error on the next inference, assuming the same depth distribution, to avoid too low or too high equivalent ground-truth. We chose the space displacement as: 
\begin{equation}
D_{optimal}(t+1) = \frac{E(\zeta(t))}{\alpha\beta_{mean}}
\end{equation}

With $E(\zeta(t))$ the mean of depth values and $\beta_{mean}$ the optimal mean output of $\beta$, e.g. $0.5$. $\Delta(t)$ is then computed numerically to get the frame shift with the closest corresponding displacement possible.

\subsection{Multiple shifts inference}

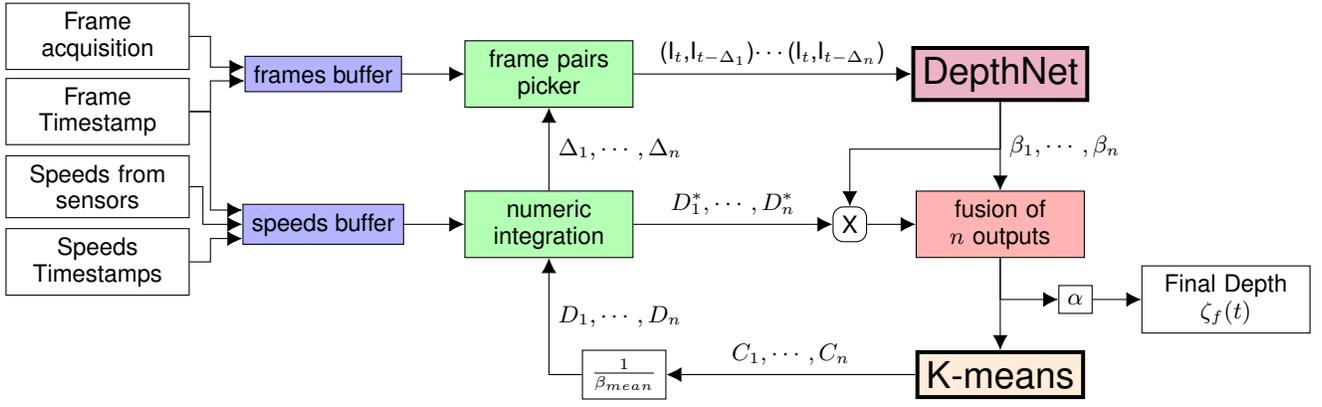
\begin{figure*}\centering
\tikzset{%
  >={Latex[width=2mm,length=2mm]},
  % Specifications for style of nodes:
            base/.style = {rectangle,draw=black,
                           minimum width=0.1cm,align=center},
            buffer/.style = {base, fill=blue!30},
            fusion/.style = {base, fill=red!30},
            integration/.style = {base, fill=green!30},
            kmeans/.style = {base, fill=orange!15},
            network/.style = {base, fill=purple!30}
}

\begin{tikzpicture}[font=\sffamily\small]
  % Specification of nodes (position, etc.)
  \node (img_input)             [base,text width=2.2cm]              {Frame acquisition};
  \node (ts_input)      [base, below of=img_input,text width=2.2cm]          {Frame Timestamp};
  \node (speeds_input)      [base, below of=ts_input,text width=2.2cm]   {Speeds from sensors};
  \node (speedts_input)      [base, below of=speeds_input,text width=2.2cm]   {Speeds Timestamps};
 
  \node (frame_buffer)    [buffer, right of=img_input, xshift=2cm,yshift=-0.5cm]   {frames buffer};
  \node (speeds_buffer)   [buffer, right of=speeds_input, xshift=2cm, yshift=-0.5cm]  {speeds buffer};
  \node (pair_picker) [integration, right of=frame_buffer,text width=2cm, xshift=2cm]   {frame pairs picker};
  \node (num_int)    [integration, right of=speeds_buffer,text width=2cm, xshift=2cm]   {numeric integration};
  \node (depthnet)      [network, right of=pair_picker, xshift=5cm, line width=1.5]   {\Large DepthNet};
  \node (fusion)    [fusion, below of=depthnet,text width=2cm, yshift = -1cm]   {fusion of $n$ outputs};
  \node (end)    [base, right of=fusion,text width=2cm, xshift=2cm, yshift = -1cm]   {Final Depth $\zeta_f(t)$};
  \node (kmeans)      [kmeans, below of=fusion, yshift = -1cm, line width=1.5]   {\Large K-means};
  \node (mult1)    [base, left of=kmeans, xshift=-4cm]   {$\frac{1}{\beta_{mean}}$};
  \node (mult2)    [base, left of=fusion, rounded corners,xshift = -1cm] {X};
  \node (mult3)    [base, left of=end, xshift=-1cm]   {$\alpha$};  
  \draw[->]     (img_input) -- +(1.5,0) |- (frame_buffer.175);
  \draw[->]     (ts_input) -- +(1.5,0) |- (frame_buffer.185);
  \draw[->]     (ts_input) -- +(1.5,0) |- (speeds_buffer.170);
  
  \draw[->]     (speeds_input) -- +(1.4,0) |- (speeds_buffer.180);
  \draw[->]     (speedts_input) -- +(1.5,0) |- (speeds_buffer.190);
  \draw[->]     (frame_buffer) -- (pair_picker);
  \draw[->]     (pair_picker) -- (depthnet)node [midway,above]{(I$_t$,I$_{t-\Delta_1}$)$\cdots$(I$_t$,I$_{t-\Delta_n}$)};
  \draw[->]     (fusion) -- (kmeans);
  \draw[->]     (fusion) |- (mult3);
  \draw[->]     (mult3) -- (end);
  \draw[->]     (kmeans) -- (mult1)node [midway,above]{$C_1,\cdots,C_n$};
  \draw[->]     (mult1) -| (num_int)node [near end,right] {$D_1,\cdots,D_n$};
  \draw[->]     (speeds_buffer) -- (num_int);
  \draw[->]     (num_int) -- (pair_picker)node [midway,right] {$\Delta_1,\cdots,\Delta_n$};
  \draw[->]     (num_int) -- (mult2)node [midway,above] {$D^*_1,\cdots,D^*_n$};
  \draw[->]     (depthnet) -- +(0,-1) -| (mult2);
  \draw[->]     (depthnet) -- (fusion)node [midway,right] {$\beta_1,\cdots,\beta_n$};
  \draw[->]     (mult2) -- (fusion);

  \end{tikzpicture}
\linebreak
\caption{Multiple shifts architecture. We used $n$ different planes. Numeric integration, given a desired displacement $D$ gives the closest possible displacement between frames $D^*$ , along with corresponding shift $\Delta$. As discussed in part \ref{uav}, the fusion block computes pixel-wise weights from $\beta_1,\cdots,\beta_n$ to make a weighted mean of $\beta_1D_1,\cdots,\beta_nD_n$}
\label{multishifts}
\end{figure*}

As neural network are traditionally computed within massively parallel architectures such as GPUs, multiple depth maps can be computed efficiently at the same time in a batch, especially for low resolution. Batch inference can then be used to compute depth with multiple shifts $\Delta(t,i)$. These multiple depth maps can then be combined to construct a higher quality depth map, with high precision for both long and short range. We propose a dynamic range algorithm, described Fig~\ref{multishifts} to compute an combine different depth maps.

Instead of only one optimal displacement $D(t)$ from $E(\zeta)$, we use K-mean clustering algorithm \cite{macqueen1967} on the depth map to find a list of clusters on which each shift will focus. The clustering outputs a list of $n$ centroids $C_i(\zeta)$ and corresponding $D_i(t)$ and $\Delta(t,i)$. $n$ is an arbitrary chosen value, usually ranging from $1$ to $4$.

Final DepthMap is then computed from fusing these outputs using a weighted mean for each pixel. Each weight is actually a linear interpolation from $0$ to $1$ according to distance of depth from a target value $\beta_{mean}$. That way, fusion will favor values that are closer to this optimal value. An $\epsilon$ value is added to solve fusion when every depth map is off its wanted range.

\begin{figure*}\centering
\includegraphics[scale = 0.5]{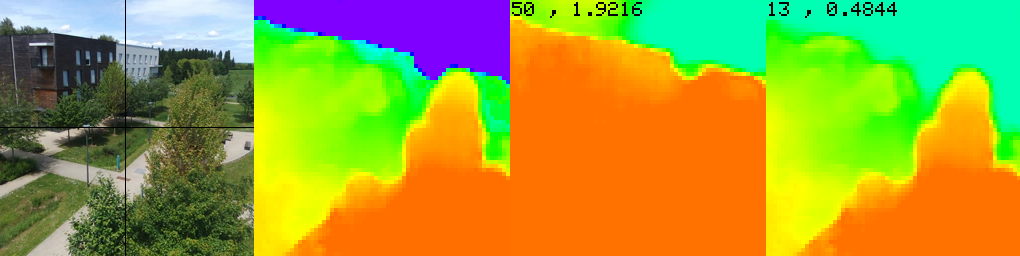}
\caption{real condition application of the multi-shift algorithm with Tiny DepthNet Clamped. First image is input. Last two are outputs of the network, for shifts of $50$ and $13$ with a drone flying forward at $1m.s^{-1}$ and at an altitude of $~12m$, with corresponding displacements from sensors. Second is fused output, capped to $100m$ up}
\label{irlex}
\end{figure*}

\begin{equation}
\begin{array}{l}
w_{ijk} = \epsilon + f(\beta(I_t,I_{t-\Delta(t,i)}))\\
f:x \mapsto \left\lbrace \begin{array}{ll}
0 &\text{ if } x<\beta_{min} \\
\frac{x-\beta_{min}}{\beta_{mean}-\beta_{min}} &\text{ if } \beta_{min} \leq x < \beta_{mean}\\
\frac{\beta_{max} - x}{\beta_{max}-\beta_{mean}} &\text{ if } \beta_{mean} \leq x < \beta_{max}\\
0 &\text{ if } x\geq\beta_{max}
\end{array}\right.\\
\zeta_i(t) = \alpha D_i(t)\beta(I_t,I_{t-\Delta(t,i)})
\end{array}
\end{equation}
\begin{equation}
\forall (j,k)\in  \llbracket0,W \rrbracket \times \llbracket 0,H \rrbracket,\zeta_f(t)_{jk} = \frac{\sum_i w_{ijk}\zeta_{ijk}(t)}{\sum_i w_{ijk}}
\end{equation}

For our use-case, we set $\beta_{min} = 0.1$ , $\beta_{mean} = 0.4$, $\beta_{max} = 0.9$ and $\epsilon = 10^{-3}$. $i$ is the index of frame shift, $j,k$ are the spatial indices. Fig~\ref{irlex} shows a result of the proposed algorithm for a batch size of $2$. Notice how the high shift detects buildings while low shift detects trees.

\subsection{Clamped DepthNet}
Our proposed algorithm is actually suffering a problem for real condition videos, because we assume a perfect stabilization. Therefore, on very far objects (e.g. the sky), any minor optical flow caused by a default in stabilization will result in a massive error in depth. Moreover, our network being very good at recognizing shapes and giving it the same depth everywhere, this can result in the whole sky being computed as relatively close.
We thus propose a network designed for a simpler problem: during training on still box, we clamp depth from $10m$ to $60m$, with a shift of $5$ images (instead of $3$ for DepthNet). These new parameters allow the network to only focus on mid range objects, dismissing close and far objects with respectively a too large and too small optical flow. This training workflow is very well suited for multiple shift depth inference. Every image pair will have a dedicated depth to analyze, allowing the fusion to not be bothered with redundant data, because of the high initial range of DepthNet.

\begin{figure}\centering
\includegraphics[scale=0.37]{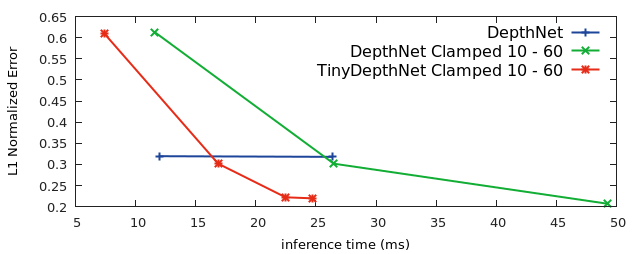}
\caption{results for synthetic $256$x$256$ scenes with noisy orientation. DepthNet has been tested with 1 and 2 planes, DepthNet Clamped with 1 to 3 planes and Tiny DepthNet Clamped with 1 to 4 planes. Y axis is Absolute mean error (m) divided by ground-truth depth, X axis is inference speed, in ms}
\label{resnoise}
\end{figure}

Figure~\ref{resnoise} shows results for multiples synthetic $256$x$256$ scenes with ground truth, along with inference speed and a small noise added to camera initial orientation at each frame. $R(t) = R_0 + Euler(N_0\mu(t))$, with $\mu(t)$ being a 3-dimensional random unit vector and $N_0$ a constant fixed to 0.001. We also report performance a thin version of our clamped network, that shows better results than DepthNet with 1 plane only in this noisy setup. The thin network has the same depth, but every convolution has an output half the number of feature maps of the original DepthNet. These results have been obtained on a Quadro K2200m powered laptop.

\section{Conclusion and future work}
We proposed a novel way of computing dense depth maps from motion, along with a very comprehensive dataset for stabilized footage analysis and a technique for dynamic range real flight computing. This algorithm can then be used for depth-based sense and avoid algorithms in a very flexible way, in order to cover all kinds of path planning, from collision avoidance to long range obstacle bypassing.

A more thorough presentation of the results can be viewed in this video. \url{http://perso.ensta-paristech.fr/~manzaner/Download/ECMR2017/DepthNetResults.mp4}

Future works include implementation of such a path planning algorithm, and construction of a real condition fine tuning dataset, using UAVs footages and a preliminary thorough 3D offline scan. This would allow us to measure quantitative quality of our network for real footages and not only subjective as for now. We could also use unsupervised techniques, using re-projection errors as in \cite{zhou2017unsupervised}.

We also believe that our network can be extended to reinforcement learning applications that will potentially result in a complete end-to-end sense and avoid neural network for monocular cameras.

The major drawback of our algorithm is however the necessity for a scene to be rigid. This is obviously never the case, and even though UAV footage are less prone to moving objects like in autonomous driving problems, we will have this issue whenever a moving target is to be followed. To solve this problem, an explicit movement equation for both the camera and the moving targets may have to be computed, as in \cite{2017arXiv170407804V}. In any case, this problem will be a challenge and may not be solvable with fully Convolutional networks only as we did in this article.

\bibliographystyle{plain}
\bibliography{biblio}

\end{document}